%% file: preamble.tex
\def\year{2022}\relax
\documentclass[letterpaper]{article} % DO NOT CHANGE THIS
\usepackage{aaai22}  % DO NOT CHANGE THIS
\usepackage{times}  % DO NOT CHANGE THIS
\usepackage{helvet}  % DO NOT CHANGE THIS
\usepackage{courier}  % DO NOT CHANGE THIS
\usepackage[hyphens]{url}  % DO NOT CHANGE THIS
\usepackage{graphicx} % DO NOT CHANGE THIS
\usepackage{natbib}  % DO NOT CHANGE THIS AND DO NOT ADD ANY OPTIONS TO IT
\usepackage{caption} % DO NOT CHANGE THIS AND DO NOT ADD ANY OPTIONS TO IT
\DeclareCaptionStyle{ruled}{labelfont=normalfont,labelsep=colon,strut=off} % DO NOT CHANGE THIS
\usepackage{algorithm}
\usepackage{algorithmic}

\usepackage{subfiles}
\usepackage{newfloat}
\usepackage{listings}
\floatstyle{ruled}
\newfloat{listing}{tb}{lst}{}
\floatname{listing}{Listing}
\usepackage{amsthm}

\theoremstyle{definition}

\usepackage{amsmath}
\usepackage{multirow}
\usepackage{graphicx}

\usepackage{fancyhdr}

\title{Recognising the importance of preference change: A call for a coordinated multidisciplinary research effort in the age of AI}

\author{
\textsuperscript{\rm 1}Matija Franklin, \textsuperscript{\rm 1}Hal Ashton, \textsuperscript{\rm 2}Rebecca Gorman, \textsuperscript{\rm 3}Stuart Armstrong}
\affiliations{
    \textsuperscript{\rm 1}University College London, \textsuperscript{\rm 2}Aligned AI, \textsuperscript{\rm 3}University of Oxford\\
    matija.franklin@ucl.ac.uk}
\begin{document}
\maketitle

\input{main}
\bibliography{preference_ref}
\end{document}

%% file: main.tex
\begin{abstract}
As artificial intelligence becomes more powerful and a ubiquitous presence in daily life, it is imperative to understand and manage the impact of AI systems on our lives and decisions. Modern ML systems often change user behavior (e.g. personalized recommender systems learn user preferences to deliver recommendations that change online behavior). An externality of behavior change is preference change. This article argues for the establishment of a multidisciplinary endeavor focused on understanding how AI systems change preference: Preference Science. We operationalize preference to incorporate concepts from various disciplines, outlining the importance of meta-preferences and preference-change preferences, and proposing a preliminary framework for how preferences change. We draw a distinction between preference change, permissible preference change, and outright preference manipulation. A diversity of disciplines contribute unique insights to this framework.
\end{abstract}

\section{Introduction}

Modern Artificial Intelligence (AI) often uses Machine Learning methods that can learn its users' preferences (e.g., favorite artist) to personalize a service to them (e.g., song recommendation; \cite{domshlak2011preferences}). User preference on online platforms is typically inferred from consumer behavior (e.g., what they have listened to) or consumer ratings (e.g., did the user click on the "like" or "dislike" button) \cite{ibrahima2018recommender,yakhchi2021learning}. The tailored recommendations change users' online behavior \cite{jesse2021digital}. Behavior change practitioners use \textit{behavioral insights} - cause and effect understanding of how factors change behavior at the level of a population - to create changes in \textit{choice architecture} - the background of people's behaviors - to influence behavior \cite{ruggeri_behavioral_2018}. Research in Behavioral science - the discipline concerned with the study of behavioral insights - has led to a valid and reliable understanding of behavior and how it changes, allowing for the development of accurate predictive models \cite{michie2011behaviour}.

It is evident that user interactions with AI-powered systems whose design has been informed by Behavioral Science lead to behavioral change. What is less evident is that these practices also cause a preference change. Preference can and does influence our behavior, but our behavior often predates and leads to the emergence of new preference \cite{ariely_how_2008}. This is an issue given that behavior change practitioners ethically defend the practice by arguing that they merely influence behavior, without limiting or forcing options \cite{sunstein_ethics_2016}. Tailoring to user preference with AI has been defended with the intent to learn about a platform's customers in order to improve their user experience. Due to the present preference change, we argue that the practices in AI and Behavioral Science can be much more manipulative \cite{sunstein2015fifty}. It is thus impossible to ensure the ethical and safe use of AI without an understanding of the impact these technologies have on preference. Furthermore, any attempt to align AI so that its "objective is to maximise the realization of human preferences" \cite{russell2019human} is futile without the acknowledgment that AI can change human preference. 

To this end, this article argues for the establishment of a multidisciplinary endeavor focused on the operationalization of preference in the broader sense, and an understanding of how it changes. In this article, we will refer to this endeavor as \textit{Preference Science} for brevity. It seeks to tackle issues central to establishing this endeavor. First, it seeks to operationalize preference in a way that incorporates features of the concept from various disciplines in order to broaden our understanding of what it is. Second, it considers \textit{meta preference} - people's preferences over what their preferences are or shall be in the future - and \textit{preference-change preference} - preferences about how preferences are formed. Third, it proposes a framework for understanding how preferences change. Finally, the article will propose a research direction for science that seeks to understand how preferences change.

\section{Preference redux}
Social sciences have operationalized and studied preference, and related concepts, in ways that are specific to those disciplines\footnote{In research design, operationalization is the process of defining a concept which cannot be directly measured}. Each of these disciplines illuminate on important properties of preferences. A definition of preference must account for this, to arrive at a more robust operationalization of preference.

We propose preference be defined as \emph{any explicit, conscious, and reflective or implicit, unconscious, and automatic mental process that brings about a sense of liking or disliking for something}. This definition broadens the scope of preference, allowing for the inclusion of other concepts that pertain to conative psychological processes. \textit{Conative mental states} are those which aim to bring about something in the environment, rather than describe it. Conative mental states inform us about what the world ought to be like, in contrast to \textit{cognitive} mental states, which inform us about what the world is like \cite{schulz2015preferences}.

\section{Preferences and conative states}

In economics, preference is often operationalized as a choice between alternative options. A choice of one option over another indicates a preference towards it. This is an example of a `revealed preference' \cite{samuelson1938note} -- the preference is assumed to be revealed through the agent's actions. It is clear that this is but a partial model of preference, as humans are not fully rational calculating machines with infinite time and knowledge. Approaches have been developed to extract preferences that acknowledge that humans are 'boundedly rational' \cite{boundedrat} - that is to say mainly rational but with certain imperfections. \citet{armstrong2018occam} formally demonstrate that, unless one makes normative assumptions, one cannot learn an agent's level of rationality \emph{and} their preferences from observation.

`Stated preferences' \cite{kroes1988stated} are often used in philosophy and in some areas of social science: the preferences of an agent are elicited by directly asking them. This has problems; people may not honestly report their true preferences \cite{sunstein2018unleashed}, and they may even be influenced in how the questions are phrased \cite{thaler2004save,vspecian2019precarious}.

Psychological research operationalizes preferences both in terms of a decision, or in terms of a judgment towards an object; specifically, liking or disliking it \cite{kahneman1982psychology}. The field acknowledges that preferences can be influenced by a person's physical and social environment, as well as their previous behavior and life history \cite{michie2011behaviour, cialdini1987influence}.

Psychologists make a conceptual distinction between preference and desire \cite{pettit1998desire, schulz2015preferences}. Desire is directed at a single object, while preferences relate to a comparison between alternatives. A preference can thus be operationalized as a comparison between desires. Each desire towards an object will have a level of intensity. People display different judgments for the same object when making \textit{relative} judgments between alternatives versus stating an \textit{absolute} judgment towards a particular object \cite{bazerman1992reversals, azar2011people}. Both conative states are thus relevant for a preference science.

\section{Preferences in machine learning}

In industry today, user preferences are heavily studied by machine learning practitioners developing personalized recommender systems. Personalized recommender engines seek to model a user's preferences in order to provide users with a tailored selection of options on a certain platform. To do this, they use data on the user's previous behavior -- this is thus a form of revealed preference\footnote{Inverse reinforcement learning makes this assumption explicit, estimating user's `reward function' under the assumption that the user is rational \cite{ng2000algorithms} or noisily rational \cite{Ziebart}.}. Systems that include some of the user's stated preferences, looking at their rating of pieces of content on the platform, have been widely deprecated due to a lower level of effectiveness at predicting user behavior. Collaborative filtering, which has been found to be effective in recommender systems, assumes users' preferences are similar to those of other users with similar behavior or background.

Commercial recommender engines are designed with the platform's interest in mind, and hence target the user's behavior - aiming to get them to click through, to buy, to spend more time on the platform, etc\ldots. By nudging a user to consume certain information on the website, recommender engines, through the bidirectional causal relationship between preference and behavior, can cause a runaway feedback-loop which over time changes the user's preference \cite{evans2021user}. \citet{alfano2020technologically} has found that such a system will recommend extremist content to maintain user engagement, creating internet echo chambers and polarization on social media platforms. \citet{Ashtonfranklin_forthcoming} discuss the general problem of AI systems manipulating user preferences in greater depth.

\section{Preferences in behavioral science}

In behavioral science, notable frameworks of behavior change have identified conative mental states that can be used to broaden our understanding of preferences. The frameworks also lay out the requirements for establishing a framework of preference change.

\textit{The COM-B model} argues that for any behavior there are three components: Capability, Opportunity, and Motivation \cite{michie2011behaviour}. To perform a behavior, a person must feel they are able to, have the opportunity for the behavior, and want or need to carry out that particular behavior (more so than any other behavior). The three components interact, and behavior change occurs as a result. An extension to this model is the \textit{Behavior Change Wheel} which describes how the components of COM-B get influenced by \textit{Choice Architecture} - the omnipresent, and influential contexts in which people behave \cite{michie2014behaviour}

Capability either relates to a person's\textit{ psychological} or \textit{physical} skills. Opportunity relates to either the opportunity afforded by the \textit{social environment}, including social cues, cultural norms, social expectations, and other people's behavior, or by the \textit{physical environment}, involving time, resources, and location. Motivation describes the conative psychological processes that are involved in behavior, and are thus applicable to our broader definition of preference. It is a mental state which relates to people's initiation, continuation, and termination of certain behaviors at a particular time. Motivation, in other words, brings about an explicit or implicit liking or disliking for behaviors. COM-B uses a dual-process approach to recognize that motivation can occur due to both automatic and reflective processes. \textit{Reflective motivation} includes conscious mental processes including attitudes and beliefs, while \textit{automatic motivation} is an umbrella term for less conscious mental processes including desires, emotional reactions, and habits. 

The \textit{Theoretical Domains Framework} (TDF) – a list of factors that influence behavior – identifies corresponding conative states that map onto the components of COM-B’s reflexive and automatic motivation \cite{atkins2017guide}. Here \textit{reinforcement} (i.e., an automatic response to a stimulus) and \textit{emotion} map onto automatic motivation. \textit{Social/professional role and identity}, \textit{beliefs about capabilities and consequences}, \textit{intentions} and \textit{goals} map onto reflexive motivation.

Three things become evident from the behavioral science literature. First, concepts that have not been traditionally thought of as preferences can be as they are conative states that bring about a sense of liking or disliking for something. Second, these conative states are not only conscious but also often automatic. Finally, preferences are shaped by multiple factors, thus, an individual's preferences will often change. Preference science must account for these factors, as well as for people's preferences towards their own preferences and how they change.

\section{Meta-preference}
People's preferences can conflict with each other -- self-regulation can be seen as successfully privileging long-term goals over short-term pleasure \cite{doerr2010self}. People will have preferences over which of these should win out, and at what cost; these preferences over preferences can be termed meta or second-order preferences.

Humans can endorse or un-endorse some of their own preferences \cite{frankfurt1988freedom}: for instance, people can harbor sexist or racist instincts which they desire not to have. And they can want to be a specific sort of person - even at a cost to themselves \cite{hewitt1990perfectionism}.

Examples of preference conflict and resolution can happen when a subject is confronted with a novel situation or thought experiment beyond their normal experience (consider the trolley problem \cite{thomson1976killing} or the repugnant conclusion \cite{parfit1984reasons}). People often resolve these conflicts in a contingent and non-consistent way, depending on the details of the circumstances they're in, their social environment, and how the issue is presented \cite{payne1993adaptive,schuman1981context,kleiman2017learning}. But they also seem to have a desire for consistency in their ethics \cite{schuman1996questions,kleiman2017learning}. This desire for consistency can itself be seen as a meta-meta or third-order preference.

Many fields define some form of idealized preferences \cite{arneson1989equality,sunstein2003libertarian,rosati2009relational}, and claim that truly serving a person's interest is to respect these idealized preferences. These idealized preferences are typically those that the person would have with, for instance, rational reflection, full information, no pressure, and enough time to ponder.

But those requirements are themselves meta-preferences, that determine the idealized preferences. It is very possible, for instance, for humans to feel that their true preferences emerge in the more emotional and dynamic parts of their lives \cite{bazerman1998negotiating,caplan2001rational}. Indeed, emotional decision-making can be more consistent than non-emotional \cite{lee2009search}. Thus the very definition of idealized preferences -- the definition of what it is to act in someone's best interests -- depends on taking their meta-preferences into account\footnote{Though these must be balanced against standard `first-order' preferences as well; a weak meta-preference should not automatically override a strongly held first-order preference.}.

We believe meta-preferences are a necessary component for building any framework concerning the ethics of preference change.

\section{Preference-Change Preferences}
As well as meta-preferences, we must also recognise preferences concerning the method of the formation process of any preference. Even if preferences meet the meta-preference requirements, how they get there is important; the ends do not always justify the means. This is recognised in the Article 5 of the draft EU AI Act \cite{cnect_proposal_2021}, which prohibits the use of any AI system that \textit{deploys subliminal techniques...%beyond a person’s consciousness
in order to materially distort a person’s behavior in a manner that causes...%or is likely to cause
that person or another person physical or psychological harm}. \citet{colburn2011autonomy} argues that subliminal or unconscious preference change devices are wrong because they interfere with a user's autonomy. They are preferences which the user cannot correctly understand why they have, because they were unaware of the process which caused them. With this account preference manipulation and induced addiction are not acceptable practices whilst self induced processes like learning and character planning are.

Guidance concerning what behavioral change mechanisms are and are not acceptable can be considered preference-change preferences given the causal relationship between behavior and preference. One guidance advocates for "a right not to be manipulated" \cite{sunstein2021manipulation}. Behavior change is said to be manipulative if it does not engage with people's capacity for reflective choice \cite{sunstein2015fifty}. Other guidance argues for "Nudge, not sludge" -- making welfare-promoting behaviors easier to do, and removing \textit{sludge} -- behavior change strategies that have the behavior change practitioner's best interest in mind rather than that of the target individual \cite{thaler2018nudge}. Sludge is said to take two forms: discouraging a person's best interest or encouraging self-defeating behaviors. A common mechanism used in sludge is friction -- making a behavior slower to do or unnecessarily complicated \cite{sunstein2018sludge}. Anti sludge guidance proposes for sludge audits -- identifying the mechanisms that change behavior and preference, and eliminating those not in line with our preference-change preferences \cite{sunstein2020sludge}. A final guidance comes from \textit{Libertarian Paternalism}. Proponents of it argue that behavior change is ethical when it avoids material incentives and coercion, and it is used to give people a guidance for best practice given their own goals, rather than changing their end goals \cite{thaler_libertarian_2003}. Altogether, there is a strong reason to have a preference-change preference for preference change mechanisms which are not manipulative, have one's best interest in mind, and aid towards one's goals.

\section{Preference change: A framework}
Preference science needs a framework that identifies the factors that influence preference and defines the causal relationship between these factors. The proposed framework is not a comprehensive list of factors, but rather a statement about how these factors group into larger ones, and relate to each other to bring about changes in preferences. Our proposed framework is visually represented in Figure 1.

\begin{figure}
\begin{center}
 \includegraphics[width=.45\textwidth]{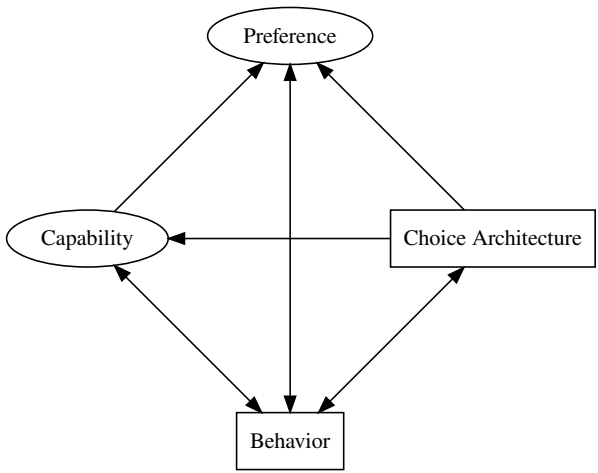}
\end{center}
\caption{Preference Change Framework: Causal diagram showing the relationship between preference, behavior, capability and choice architecture}
\label{fig:pentagram}
\end{figure}

The framework outlines how behavior, capability, and choice architecture shape preference. Preference is operationalized with the aforementioned definition, thus including all conative mental states, both reflective and automatic. In the framework, preference is influenced by choice architecture, in that conative states by definition get influenced by what is present in the environment \cite{michie2014behaviour}. Preferences are also influenced by capability, in that people will on average prefer activities (as well as objects associated with those activities) that are easier for them to do \cite{lee2011research, juvina2018measuring}. Finally, preference is in a bidirectional causal relationship (i.e., feedback loop) with behavior \cite{albarracin2005maintenance, ariely_how_2008, wyer2012effects, hill2019choice}. Preference shapes behavior, but behavior often predates and causes the emergence of new preferences. 

Behavior is the activity of a human individual, interacting with their environment \cite{popescu2014human}. The way it is operationalized is context-specific. Aside from behavior's relationship with preference (see above), behavior is in a bidirectional causal relationship with capability and choice architecture. Evident in research showing how practice and learning result in skill development, behavior influences capability \cite{billett2010learning, vaci2019joint}. Evidence for capability's influence on behavior comes from research showing that people tend to avoid activities that they find physically or psychologically difficult \cite{kool2010decision, feghhi2021effort}. Behavior change practitioners use this insight - if you want to increase a particular behavior in a population make it easier for people to do \cite{halpern2015inside}. Finally, behavioral science research has well documented the effects of choice architecture on behavior \cite{michie2011behaviour, thaler2021nudge}, and individuals behaving in an environment will cause changes to the environment (i.e., choice architecture).

As with the COM-B framework \cite{michie2011behaviour}, capability refers to people's psychological or physical skills. Capability is operationalized in terms of changes in performance. Psychological skills include, but are not limited to, knowledge, cognitive skills, memory, interpersonal skills, attention, behavioral regulation, and willpower \cite{atkins2017guide}. Physical skills refer to various abilities for different activities, as well as strength and stamina. Aside from capability's previously discussed relationship with preference and behavior, capability is causally influenced by choice architecture. Certain environments promote more (skill) learning than others \cite{land2012theoretical, gilavand2016investigating, amundrud2021advancing}. In behavior change, these insights are captured with \textit{boosts} -- behavioral interventions which promote people's capabilities \cite{franklin2019optimising, hertwig2017nudging}.

Choice architecture, for the present framework, is conceptualized with its original definition, proposed by \cite{thalernudge}. Choice architecture is the environment in which people behave \cite{thaler2013choice}. The use of choice architecture in the present framework differs from how it is used in COM-B, in that it captures both opportunity and choice architecture as described in the later framework \cite{michie2011behaviour}. Choice architecture in the present framework thus captures the resources needed for a behavior afforded by the physical (e.g., time, location) and social (e.g., language) environment. It also captures aspects of the physical and social environment that influence changes in capability, behavior, and preference. Not all aspects of the environment are equally influential, with different aspects having a larger or smaller effect size. There have been several documented aspects of the physical environment being influential with a full review being outside of the scope of the present paper \cite{ruggeri_behavioral_2018, thaler2021nudge}. Influence from the social environment broadly relates to either influence from other people's observable behavior or stated preferences \cite{cialdini2004social, cialdini2010social}. In behavioral science, these sources of influence are referred to as descriptive social norms –- how most people behave -– and injunctive social norms –- what most people prefer and value \cite{reno1993transsituational}.

\section{Future research directions}

A robust science of preference will benefit from contributions from the bodies of knowledge of diverse fields, including but not limited to psychology, cognitive science, computer science, marketing, economics, philosophy, and law. Translational research using existing insights from these fields would be beneficial. Such research could expand the scope of preference, as well as our understanding of preference change, meta-preference, and preference-change preference.

Preference science could benefit from new empirical and simulation studies. This would involve developing paradigms focusing on preference as the main dependent variable that can explore how controlled changes in capability, behavior, and choice architecture predicatively change preference. Simulation studies using valid assumptions for preference change mechanisms can also allow researchers to test hypotheses.

Internal researchers in industries that own user data and can rapidly deploy changes to user interfaces have the ability to conduct preference-science research. However, any research which could be categorized as behavioral experimentation must be accompanied by ethical and legal protocols developed for such cases, even when conducted in a commercial context. Computer Engineers must educate themselves about the ethics of user manipulation, and it should be incorporated as a core part of their formal education to help the effort in making any research conducted behind closed doors ethically and legally sound. For this to happen, Preference Science should address when preference change is ethically appropriate and legally permissible. That altering preferences of users necessarily requires ethical consideration is a major contribution of this paper. 

\section{Conclusion}

This article argues for a multidisciplinary effort to study preference change processes motivated by the observation that AI and ML system owners are altering user preferences for their own purposes. This is often brought about through behavior change techniques. We have introduced a broad definition of preferences and discussed the important issues of meta and preference-change preferences. We present a framework that looks at the dynamic interplay between choice architecture, capability, and behavior, and how this results in preference change. We argue that society needs to develop legal and ethical frameworks to identify acceptable from abusive preference change practices. 